\documentclass[a4paper, 10pt, conference]{ieeeconf}      %
\usepackage{FG2024}
\usepackage{graphicx}
\usepackage{amsmath}
\FGfinalcopy %

\usepackage[utf8]{inputenc} %
\usepackage[T1]{fontenc}    %
\usepackage{hyperref}       %
\usepackage{url}            %
\usepackage{booktabs}       %
\usepackage{amsfonts}       %
\usepackage{nicefrac}       %
\usepackage{microtype}      %
\usepackage{xcolor}         %
\let\labelindent\relax
\usepackage{enumitem}
\usepackage{textpos}
\usepackage{tcolorbox}
\tcbuselibrary{skins}
\usepackage{lipsum}
\usepackage{graphicx}
\usepackage{caption}
\usepackage{subcaption}
\usepackage{float}
\usepackage[font=small,labelfont=bf]{caption}
\usepackage{listings}
\usepackage{placeins}
\usepackage{etoolbox}
\usepackage{bm}
\usepackage{graphicx} %
\usepackage{capt-of,etoolbox}
\IEEEoverridecommandlockouts                              %
\def\FGPaperID{50} %

\title{\LARGE \bf
SignAvatar: Sign Language 3D Motion Reconstruction and Generation 
}

\author{\parbox{16cm}{\centering
    {\large Lu Dong,  Lipisha Chaudhary, Fei Xu,  Xiao Wang,  Mason Lary,  Ifeoma Nwogu}\\
    {\normalsize
     Department of Computer Science and Engineering, University at Buffalo, NY, USA}
    }
}

\makeatletter
\apptocmd{\@maketitle}{\myfigure}{}{}%
\makeatother

\usepackage{fancyhdr}
\begin{document}

\newcommand\myfigure{%
\centering
    \includegraphics[width=0.98\linewidth]{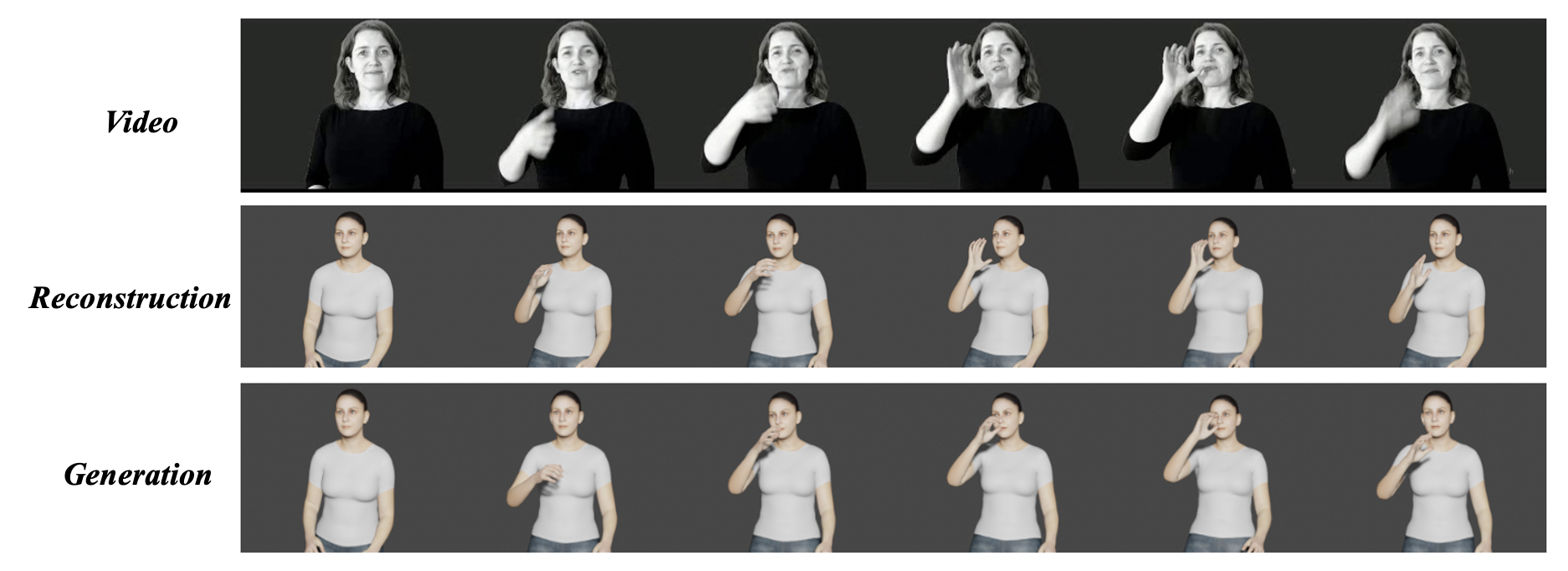}
\captionof{figure}{SignAvatar excels at two tasks: reconstructing 3D sign language motions from videos and generating them from semantics (images, text). The top row displays a sign language video for "drink" - note some motion blur here. The middle row shows the 3D avatar reconstruction by SignAvatar, and the bottom row demonstrates its ability to generate a 3D signing avatar from the word "drink".}
\label{fig:quality}
}

\ifFGfinal
\thispagestyle{empty}
\pagestyle{empty}
\else

\pagestyle{plain}
\fi

\maketitle
\thispagestyle{fancy}
\begin{abstract}

Achieving expressive 3D motion reconstruction and automatic generation for isolated sign words can be challenging, due to the lack of real-world 3D sign-word data, the complex nuances of signing motions, and the cross-modal understanding of sign language semantics. To address these challenges, we introduce SignAvatar, a framework capable of both word-level sign language reconstruction and generation. SignAvatar employs a transformer-based conditional variational autoencoder architecture, effectively establishing relationships across different semantic modalities.
Additionally, this approach incorporates a curriculum learning strategy to enhance the model’s robustness and generalization, resulting in more realistic motions. Furthermore, we contribute the ASL3DWord dataset, composed of 3D joint rotation data for the body, hands, and face, for unique sign words. We demonstrate the effectiveness of SignAvatar through extensive experiments, showcasing its superior reconstruction and automatic generation capabilities. The code and dataset are available on the project page\footnote{\href{https://dongludeeplearning.github.io/SignAvatar.html}{https://dongludeeplearning.github.io/SignAvatar.html}}.

\end{abstract}

\addtocounter{figure}{-1}

\FloatBarrier

\section{INTRODUCTION}

Over 70 million people worldwide rely on sign language as their primary way of communication \cite{davis2019hearing}. The emergence of AI has catalyzed research in Sign Language Recognition (SLR) and Sign Language Translation (SLT). Most of the works \cite{moryossef2021evaluating, miah2023multi} use skeletal data to analyze motions and learn feature projections with designed models. However, this approach may not be well-suited for generative tasks because synthetic skeletal motion is highly abstract and may not be readily understood by DHH individuals, hence is not as conducive for effective communication. The use of virtual human agents, signing avatars \cite{naert2020survey}, holds the potential for automated interaction in scenarios such as remote communication, customer service, or public announcements \cite{quandt2022attitudes}. Using 3D word-level sign dictionaries \cite{naert2020survey} can greatly improve sign language synthesis and enhance the quality of sign language learning. 
Initial effort \cite{hanke2004hamnosys} required extensive manual intervention and expensive equipment support, resulting in avatars with unnatural movements. 

Our goal in this work, therefore, is to synthesize expressive and natural 3D word-level signing avatars, either by reconstructing them from videos or automatically generating them from texts only.
Reconstructing realistic 3D sign motions from 2D videos and semantically controlling such motion generation both present significant challenges. Firstly, it requires reliable 3D data for learning. However, the equipment needed for capturing such data is expensive and requires skilled professionals to operate, which discourages many researchers.  
Secondly, unlike the broad body motions often considered in many activity recognition tasks, the motions in sign language typically involve subtle variations. Even slight changes in these gestures can lead to significant shifts in meaning.
The definitions of sign language grammar encompass various factors, including whether signs are one-handed or two-handed, the shape of gestures, palm orientation, locations, and movement patterns. Consequently, mastering these characteristics presents considerable challenges in learning and distinguishing between different signs.

To tackle the aforementioned challenges, we introduce SignAvatar, a novel approach for 3D motion reconstruction and automatic sign language generation, providing an effective solution for learning, understanding, and generating signs. 
Our SignAvatar employs a transformer-based conditional variational autoencoder (CVAE) framework, capturing temporal and spatial relationships in sign language motions while aligning semantics. %
We leverage the Contrastive Language-Image Pre-Training (CLIP) \cite{radford2021learning} latent space for conditioning. This model not only effectively captures textual semantics but also offers semantic expansion for images. As a result, our SignAvatar can generate sign language 3D motion from various semantic conditions. 
Further elaboration on the model architecture is provided in Section \ref{sec:method}.

Along with the framework design, we also introduced a curriculum learning training strategy to further enhance the robustness of generating consistent and diverse motions. This approach employs masked motion modeling, gradually raising the mask ratio during training. 
Detailed results showing the efficacy of curriculum learning are presented in Table \ref{table_evaluation}.

To further evaluate our SignAvatar framework, we constructed a 3D sign language motion dataset from the Word-Level American Sign Language (WLASL) dataset \cite{li2020word}. This dataset, which is a large collection of isolated videos, serves as a valuable resource for SLR tasks \cite{bohavcek2022sign}. However, we have also identified some limitations in its use for generative work.
Section \ref{sec:dataset} explains how the samples in the dataset were selected and refined for further processing. 
For 3D pose extraction, rather than the typical 3D skeleton-based features, we use the SMPL-X \cite{pavlakos2019expressive} format which includes joint rotation information for the body, hands, and face, making it suitable for flexible and complex sign language motions. Additionally, this pose format is conducive to generating SMPL-X meshes and rendering avatars. More information on this process is given in Section \ref{sec:posefeature}.

SignAvatar takes a stride towards automating the synthesis of realistic 3D signing avatars. We have tackled a series of challenges ranging from data acquisition and framework design to enhancing generalizability. Simultaneously, we have established a more robust baseline for conditional 3D signing word generation.
In summary, our contributions can be outlined as follows:

\begin{itemize}[leftmargin=*,noitemsep,topsep=0pt]
    \item We propose SignAvatar, a sign language generative framework that integrates a transformer-based CVAE architecture and a large vision-language model, CLIP. For the first time, we are able to reconstruct 3D sign language motions from isolated videos and also generate 3D motions from text or image prompts, thus representing a major advancement in automated sign language understanding technology.
    \item We introduce a curriculum learning strategy that gradually increases the mask ratio during training. This approach aids SignAvatar in enhancing its fine-grained gesture learning and generalization abilities, thereby facilitating the synthesis of realistic and natural signing motions.
    \item We provide a thorough evaluation of SignAvatar, showcasing its denoising capability in sign language reconstruction and superior capability in sign language motion generation. 
    Additionally, we contribute the ASL3DWord dataset, comprising word-level 3D joint rotation sequence, for 3D sign language research.
\end{itemize}

\section{Related Work}
\noindent \textbf{Sign Language Synthesis} \quad
In recent years, AI-generated content has seen remarkable advancements, significantly driving progress in automated sign language synthesis research. Notably, recent work by Saunders et al. includes the generation of 3D single-channel signs based on a gloss intermediary \cite{saunders2020progressive}, and the introduction of a sophisticated 3D multi-channel system. This system integrates both manual (hand movements) and non-manual (facial landmarks) features, employing advanced technologies such as transformers and mixture density networks \cite{saunders2021continuous}.
Other multi-channel approaches have also been explored \cite{saunders2020adversarial}. These sign language generation techniques were developed and evaluated using the RWTH PHOENIX-Weather-2014T (PHOENIX14T) \cite{koller2015continuous} German sign language dataset. 
Stoll et al. introduced the Text2Sign system \cite{stoll2020text2sign}, employing Generative Adversarial Networks (GANs) to produce sign videos.  
 Zelinka et al. developed text-to-video signs \cite{zelinka2020neural} using a skeletal model based on the Openpose \cite{cao2017realtime} framework. 
 A comprehensive review of sign language production was presented by Rastgoo et al. \cite{rastgoo2021sign}, mentioning the lack of annotated datasets is one of the major challenges in this field. 
Sign language production typically relies on seq2seq mapping, which can only \emph{recreate} a limited set of predefined sentences. However, due to the dynamic nature of real-life sentences, the scalability of this approach is severely limited. Research on sign language generation that begins at the word-level semantic level offers greater flexibility and better support for generating unseen semantic expressions. Since words form the basis of sentences and phrases, our approach focuses on word-level generation.

\begin{figure*}[t]
    \centering
    \includegraphics[width=0.90\linewidth]{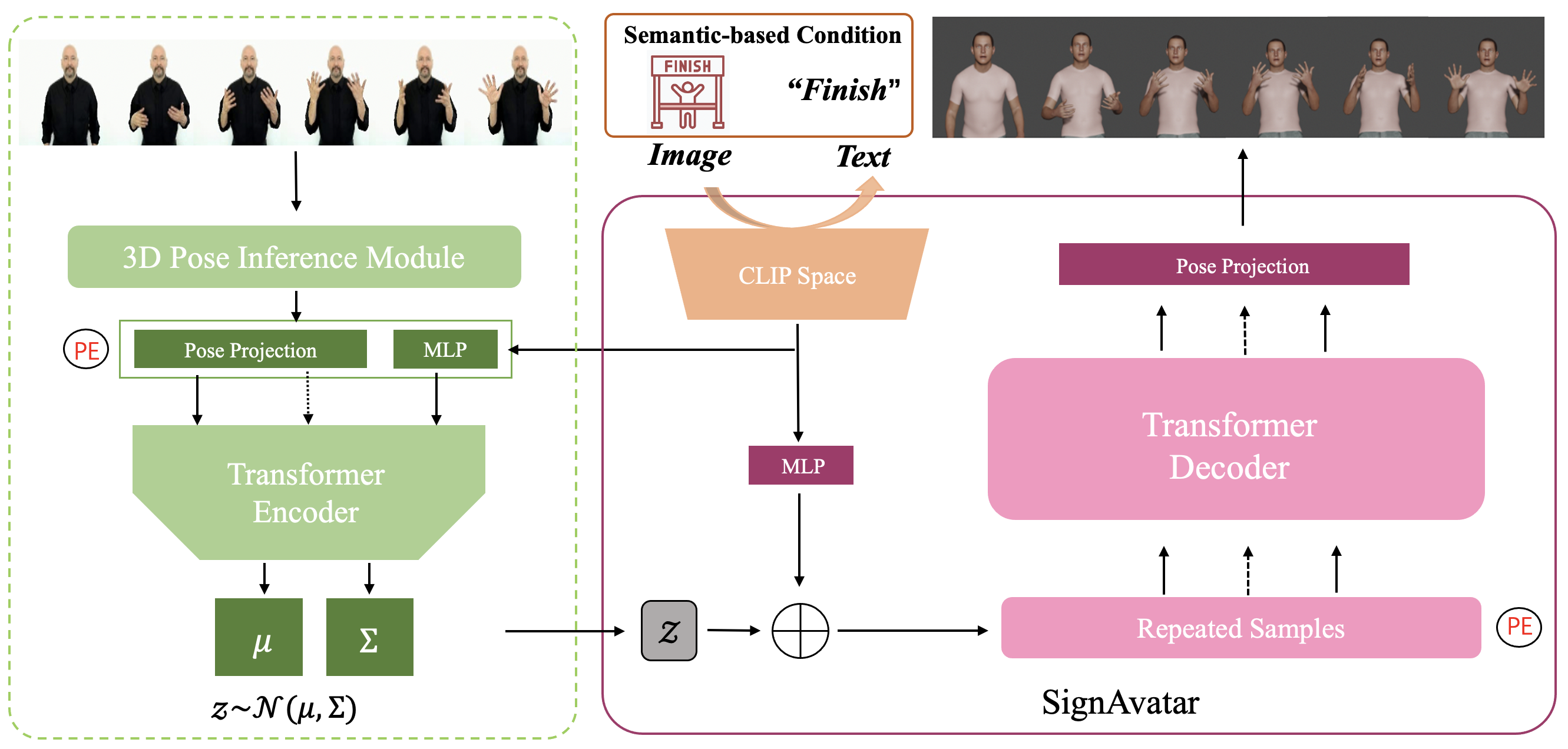}
    \caption{SignAvatar overview. We present our SignAvatar transformer-based CVAE framework, featuring an encoder (left, green dashed box) and a decoder (right, pink solid box), designed for both the reconstruction of signing avatars from videos and the generation of signing avatars from conditions (middle, yellow solid box). These conditions can be text or images, which are transformed into semantic embeddings using CLIP. Given a sign language video, the SignAvatar encoder learns the sign motion distribution through the following process: pose extraction (3D Pose Inference Module), concatenation with semantic embeddings, positional encodings (PE), and utilization of a transformer encoder. Samples 'z' from this distribution, when combined with the semantic-based condition, will guide the SignAvatar decoder in synthesizing pose sequences, which are then rendered using Blender. For reconstruction, SignAvatar samples from the learned distribution, while for generation, it bypasses the left encoder and directly samples sequences from a standard normal distribution.}
    \label{fig:framework}
\end{figure*}

\noindent \textbf{Pose Representation} \quad
Transitioning from 2D skeletons to 3D skeletons \cite{saunders2021continuous, saunders2020adversarial, zelinka2020neural} for the hands, body, and facial features has become mainstream due to its incorporation of depth, thereby enhancing our understanding of motion. However, this approach still results in unrealistic abstraction by losing shape information, which is a critical component in sign language understanding.
In addition to methods that estimate a sparse set of landmarks, various approaches have been developed for estimating parameters of morphable models for different body parts. These include the hand \cite{abs-1904-05767, Kulon_2020_CVPR, Moon_2020_ECCV_I2L-MeshNet, Zhang_2021_ICCV}, face \cite{Danecek_2022_CVPR}, and body \cite{Choi_2021_CVPR, joo2020eft, Kanazawa_2018_CVPR, Kocabas_2021_ICCV, Kolotouros_2019_ICCV, Li_2021_CVPR, Muller_2021_CVPR, Zhang_2021_ICCV}. 
The emergence of expressive 3D body models like SMPL-X \cite{Pavlakos_2019_CVPR}, Adam \cite{Joo_2018_CVPR}, and GHUM \cite{Xu_2020_CVPR} has made it easier to research and estimate complete 3D body surfaces. However, in the domain of sign language, these body models have not been fully utilized. 
We believe that leveraging SMPL-X to synthesize expressive, human-like sign language avatars represents a significant step forward in the field.

\noindent \textbf{3D Human Pose Estimation from Single Image} \quad
Several initial studies employed an optimization-based approach, wherein they aimed to fit a 3D human model to the available 2D/3D ground-truth data. 
For instance, Joo et al. \cite{Joo_2018_CVPR} employed this approach by fitting their human models to 3D human joint coordinates and point clouds, in a multi-view studio setting. Building on their work, Xiang et al. \cite{Xiang_2019_CVPR} extended the approach to single-view RGB scenarios. In parallel, SMPLifX \cite{Pavlakos_2019_CVPR} and Xu et al. \cite{Xu_2020_CVPR} also utilized optimization-based techniques, to fit their respective human models, SMPL-X and GHUM, to 2D human joint coordinates. However, these optimization-based methods are known to be slow and sensitive to noisy evidence.

More recently, a shift towards regression-based approaches has been observed. Numerous neural network-based methods have emerged, each comprising of distinct networks dedicated to body, hand, and face components. Each of these networks takes as input a human image, a hand-cropped image, and a face-cropped image, respectively, to predict SMPL-X parameters and generate a final whole-body 3D human mesh. Notable regression-based methods in this context include ExPose \cite{choutas2020monocular}, the approach that employed three separate networks. Zhou et al. \cite{Zhou_2021_CVPR} leveraged 3D joint coordinates for predicting 3D joint rotations, while PIXIE \cite{PIXIE:2021} introduced a moderator to enhance the prediction of 3D joint rotations. Using a Pose2Pose design, Hand4Whole\cite{moon2022accurate} utilized both body and hand metacarpophalangeal joint features for accurate 3D wrist rotation and smooth connection between 3D body and hands. We incorporate this last method into our proposed generative model, to alleviate the current challenge of insufficient real-life 3D data availability\footnote{Such 3D real-life data can be obtained with motion capture sensors, depth cameras, etc., but these can be expensive and resource intensive.}.

\noindent \textbf{Conditional Generation} \quad
Researchers have consistently aimed to enhance controllability in generation, whether for motion synthesis or static images. For instance, ACTOR \cite{petrovich2021action} achieved action-aware latent representation learning for human motions using a transformer-based VAE, albeit with one-hot action labels as a condition. The subsequent development of language-guided generation established a stronger link between visual representation and semantic space, offering more precise control and expanding creative possibilities. Many prior language-guided generation approaches \cite{zhang2017stackgan,li2020unicoder,nam2018text} have primarily focused on generating images.
 DALL·E ~\cite{ramesh2021zero} employed a discrete variational autoencoder to create diverse images based on text embeddings derived from GPT-3 \cite{brown2020language}. A more recent development, CLIP\cite{radford2021learning}, demonstrated remarkable capabilities by jointly learning a multi-modal vision-language embedding space. Building on the power of CLIP, StyleCLIP \cite{patashnik2021styleclip} extended StyleGAN \cite{karras2019style} into a language-driven generation model through their CLIP-guided mapper. 

 Current action generation models primarily focus on body motion, neglecting the intricate movements of hands. This oversight makes it relatively easy to train the generative model due to fewer joints and larger movement changes. However, in sign language data, the body remains relatively static, with only the hands and upper limbs exhibiting high flexibility and often subtle motions. This presents a challenge akin to fine-grained motion learning, making it more difficult to train the model effectively.

\section{Method}\label{sec:method}

\subsection{\textbf{Problem Formulation}}
The task of sign language 3D motion synthesis aims to reconstruct motion from videos and generate sign motion sequences from labels that accurately represent the given label semantics $Y$. To convey semantics accurately, it is necessary to consider factors such as gestures, upper body movements, and facial expressions. Moreover, to make sign language more realistic and naturally reflect different morphology, it is desirable to disentangle the pose and the shape. Thus, we adopt the SMPL-X body model, a unified representation model with shape parameters trained jointly for the face, hands, and body. Considering a neutral body shape, our goal then is reduced to generating sequences of pose parameters. Specifically, we use a pose inference module to estimate the SMPL-X pose parameters $p_{t}$ from the video frames $f_{t}$ and construct word text and motion pairs $\left\{ \left( Y_{word}, M_{n} \right) \right\}$, where $M_{n}=[p_{1},.... p_{T}]$ is the motion sequences. 
During training, we utilize the upper pose information as input, while during the testing phase, the goal is to synthesize a full-body motion sequence $[p_{1},.... p_{T}]$ based on the input word text.

\subsection{\textbf{Conditional VAE with CLIP latent space} }

We model the sign language synthesis process with a Conditional Variational Autoencoder (CVAE) framework, as illustrated in the Fig. \ref{fig:framework}. The proposed SignAvatar consists of a transformer-based encoder-decoder architecture. The encoder extracts the core structure from the input motion sequence, creating a concise latent representation. The decoder then combines this latent representation with the text embedding from CLIP to produce a realistic human motion sequence that matches the specified condition.

\noindent \textbf{Encoder} \quad
Our CVAE encoder takes the pose sequence and word-level text projections as input, using a transformer architecture to learn their respective latent representations and calculate Gaussian distribution parameters, denoted as $\mu$ and $\Sigma$. %
We concatenate two tokens to represent the Gaussian distribution parameters, which are derived from the input embedding using a three-layer multilayer perceptron (MLP).
For word-level embedding, we employ CLIP as the projection method since it is trained on both images and text semantics, providing robust generalization to semantic relations and image recognition. Combining information from both sources through concatenation enhances the coupling of semantic and motion data. Additionally, using an MLP projection helps align the dimensions of motion and text, facilitating better integration of information within the same latent space. The resulting encoder input is the summation with the positional encodings in the form of sinusoidal functions. We obtain the distribution parameters $\mu$ and $\Sigma$ by taking the output from the corresponding token position while ignoring the rest.

\noindent \textbf{Decoder} \quad
When provided with a latent vector $ z $, we initially introduce a conditional bias in order to integrate categorical information. This bias is unique to the words and is acquired from the word-level embedding via MLP layers. Additionally, we introduce a time dimension denoted as $T$, and this information is duplicated to create a sequence. To provide positional context, we employ sinusoidal positional encoding before feeding the data into the transformer decoder. 
In this context, the time information represented by $T$ sinusoidal positional encodings serves as the query ($Q$), while the $Z$ sequence infused with semantic information is treated as the key ($K$) and value ($V$).

The decoder subsequently generates a pose sequence using the latent space by projecting it back to the SMPL-X parameters. These parameters are then utilized in Blender with the SMPL-X add-on for rendering, resulting in the creation of realistic 3D avatars, with options for both male and female avatars. Notably, our model operates efficiently, generating the entire sequence at once, without the need for an autoregressive approach that relies on prior information to generate the next pose.

\noindent \textbf{Learning Objectives } \quad
In line with the standard VAE approach, our model was trained with two fundamental components: the reconstruction loss $\mathcal{L}_\text{rec}$ and  Kullback-Leibler (KL) divergence loss $\mathcal{L}_{\text{KL}}$. 
The primary objective of the reconstruction loss is to minimize the dissimilarity between the original motion representation $\bm{M}$ and the reconstructed motion representation $\hat{\bm{M}}$ using a mean square error loss: 
\begin{equation}
    \mathcal{L}_{\text{rec}} = \frac{1}{T}\sum_{t=1}^{T} \left \| \bm{p}_t - \hat{\bm{p}}_t \right \|_2^2. 
\end{equation}
On the other hand, the KL divergence loss $\mathcal{L}_{\text{KL}}$ minimizes the distribution difference between the estimated posterior $N(\mu, \Sigma)$ and the prior normal distribution $N(0, I)$.
Thus, the CVAE training loss $\mathcal{L}_{\text{CVAE}}$ is a weighted sum of
the two terms:
\begin{equation}
    \mathcal{L}_{\text{CVAE}} = \mathcal{L}_{\text{rec}} + \omega_{\text{KL}}\mathcal{L}_{\text{KL}},
\end{equation}
where $\omega_{\text{KL}}$ is a weighting hyperparameter.

\subsection{\textbf{Curriculum Learning Strategy}}
\begin{figure*}[t]
    \centering
    \includegraphics[width=0.70\linewidth]{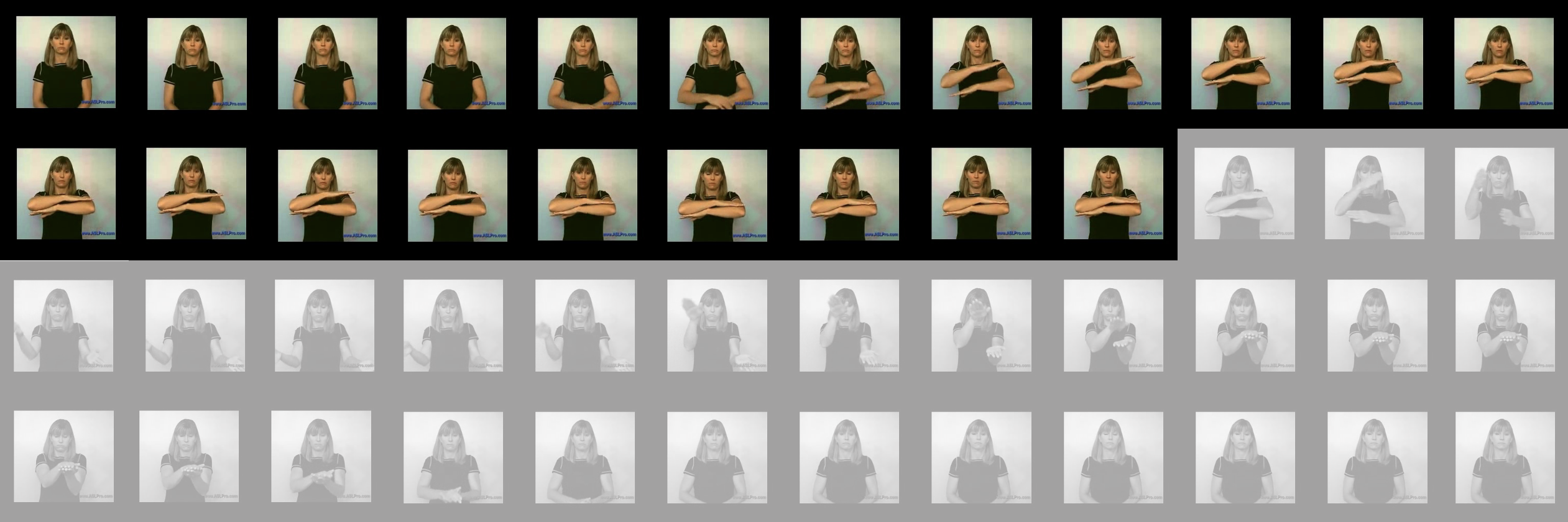}
    \caption{ Data Collection Quality Control Process. This graph displays the downsampled video frames of certain "Table" videos. The correct grammar for this sign-word involves holding both hands and forearms horizontally in front of the body, with the dominant forearm positioned above the non-dominant one, followed by tapping them together. The 21 frames shown earlier match this description, while the gray area that follows does not; these frames will be manually removed.}
    \label{fig:cleaning}
\end{figure*}
Employing curriculum learning, where models are progressively exposed to easier and then more challenging samples, has proven to enhance performance\cite{wang2021survey}. Recent work \cite{he2022masked, zhai2023language} demonstrated that even when we mask a portion of inputs, meaningful outcomes can still be achieved. This masking operation essentially elevates the sample's complexity. Inspired by the success of these works in effective representation learning, we introduce curriculum learning of masked sign motion modeling. This technique entails randomly masking a portion of the input motion sequence at a ratio $r$ and tasking the model with reconstructing the complete motion sequence. 

We perform curriculum learning by
progressively increasing the mask ratio as the training progresses according to a growth function $g(ep)$, where $ep$ represents epochs.
\begin{equation}
    {g(ep) = min\left\{  0.1\ast  \left\lfloor \frac{ep}{500} \right\rfloor, 0.6\right\}, ep\in [0,5000)}
\end{equation}
As a result, at the beginning of training, a lower mask ratio allows the model to learn
basic motion patterns and capture fundamental structures in the motion data.
As the mask ratio increases, the model is exposed to more challenging and
complex motion sequences.
This curriculum strategy thus enables a more effective and stable learning
experience, ultimately improving motion synthesis performance. It enhances the model's comprehensive grasp of motion data, encompassing both local and long-range patterns.

\section{Dataset}\label{sec:dataset}

To quantitatively evaluate SignAvatar's performance in sign language reconstruction and generation, we construct the ASL3DWord dataset from the WLASL video dataset \cite{li2020word}. WLASL is the largest publicly available Word-Level American Sign Language (ASL) dataset for SLR, containing 2000 words (glosses) and their corresponding sign videos.
However, while WLASL boasts of such a large vocabulary, its data distribution of videos per word is quite unbalanced, where there is an uneven number of sign video samples per word. This number ranges from 7 to 40 video samples per word. Having such a varied number of samples will inevitably skew the motion distribution learned and affect the correctness of generated samples.

\begin{figure}[H]
    \centering
    \includegraphics[width=0.6\linewidth]{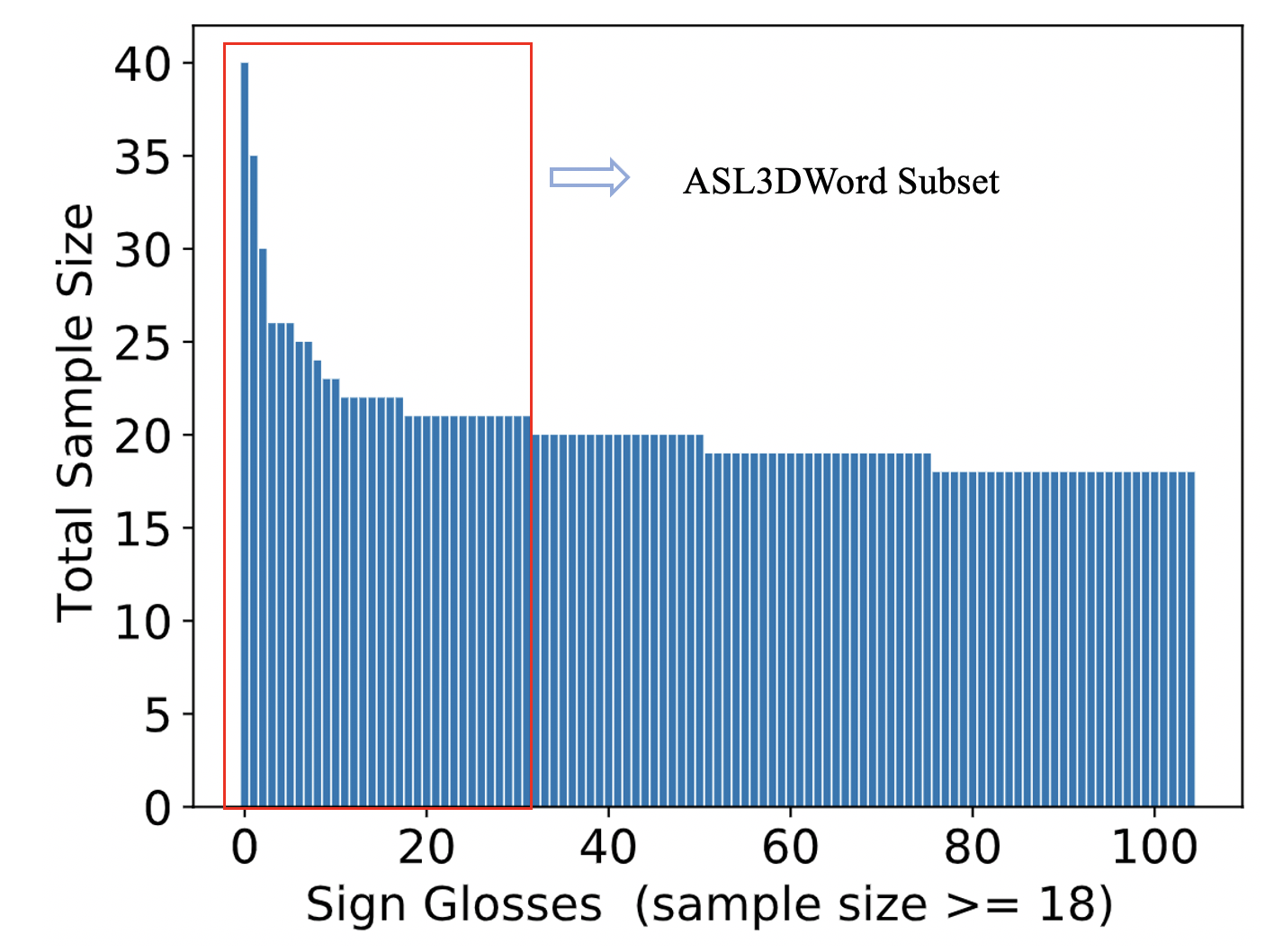}
    \caption{ {ASL Word sample statistics in WLASL: shown for words having a sample size of 18 or more videos per word. For generative models such as VAE and CVAE, a small number of samples cannot properly capture the underlying distribution for the word sign motion.} }
    \label{fig:wlasl}
\end{figure}
Thus, to create a more balanced dataset, with a sufficient number of samples needed to train each word sign, we first filtered out the words having less than 20 video samples, to build the source video subset. To create the larger dataset, from the original data, we again filtered out the words having less than 18 video samples. The resulting data distribution in our curated dataset is shown in Fig. \ref{fig:wlasl}. 
\subsection{\textbf{Quality Control} }
After some preliminary analysis, it was observed that the dataset contained considerable noise samples, where in certain video clips, only the initial portion corresponded to the intended word meaning, while the remainder contained words that had no bearing on the label of the video.
Fig. \ref{fig:cleaning} shows the 48 frames from a video featuring the semantic word "table" within the dataset. In the figure, only the first 21 frames contain the sign for the table while the rest of the frames contain other signed words which we consider as noise. Because these noise signals could negatively affect the performance of the generative model, we manually removed such "noisy" frames from the dataset.

Additionally, certain videos featured prolonged periods of inactivity at the beginning, where no signs were performed. So, to enable the model to glean as much useful information as possible, we removed excessively long silent frames, thus enhancing the overall data quality.
Ultimately, we constructed a 3D sign language dataset consisting of 103 gloss words. We then divided the data for each word into training and testing sets in an 8:2 ratio, resulting in a total of 1208 samples for training and 339 samples for testing.

\begin{figure}[H]
    \centering
    \includegraphics[width=1\linewidth]{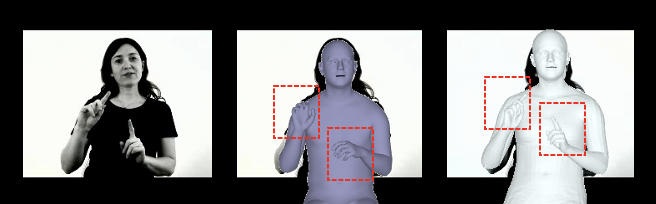}
    \caption{ Comparing 3D upper body pose estimation with different models. The original image is on the left, with ExPose\cite{choutas2020monocular} extraction in the middle and Hand4Whole\cite{moon2022accurate} extraction on the right. }
    \label{fig:upper_body}
\end{figure}

\subsection{\textbf{Pose Feature Extraction} }\label{sec:posefeature}
After completing the data quality control process, we used a 3D pose estimation model to extract 3D SMPL-X parameters from video frames. In terms of feature extraction, the optimization-based methods \cite{pavlakos2019expressive} were relatively slow, making them less suitable for extracting data in large quantities. Therefore, we focused on regression-based methods, first exploring ExPose \cite{choutas2020monocular}. It introduced body-driven attention to improve efficiency. However, we found that the results it produced for sign motion data were unsatisfactory, with unnatural wrist rotations and inaccuracies in gestures.

\begin{figure*}[t]
    \centering
    \includegraphics[width=0.86\linewidth]{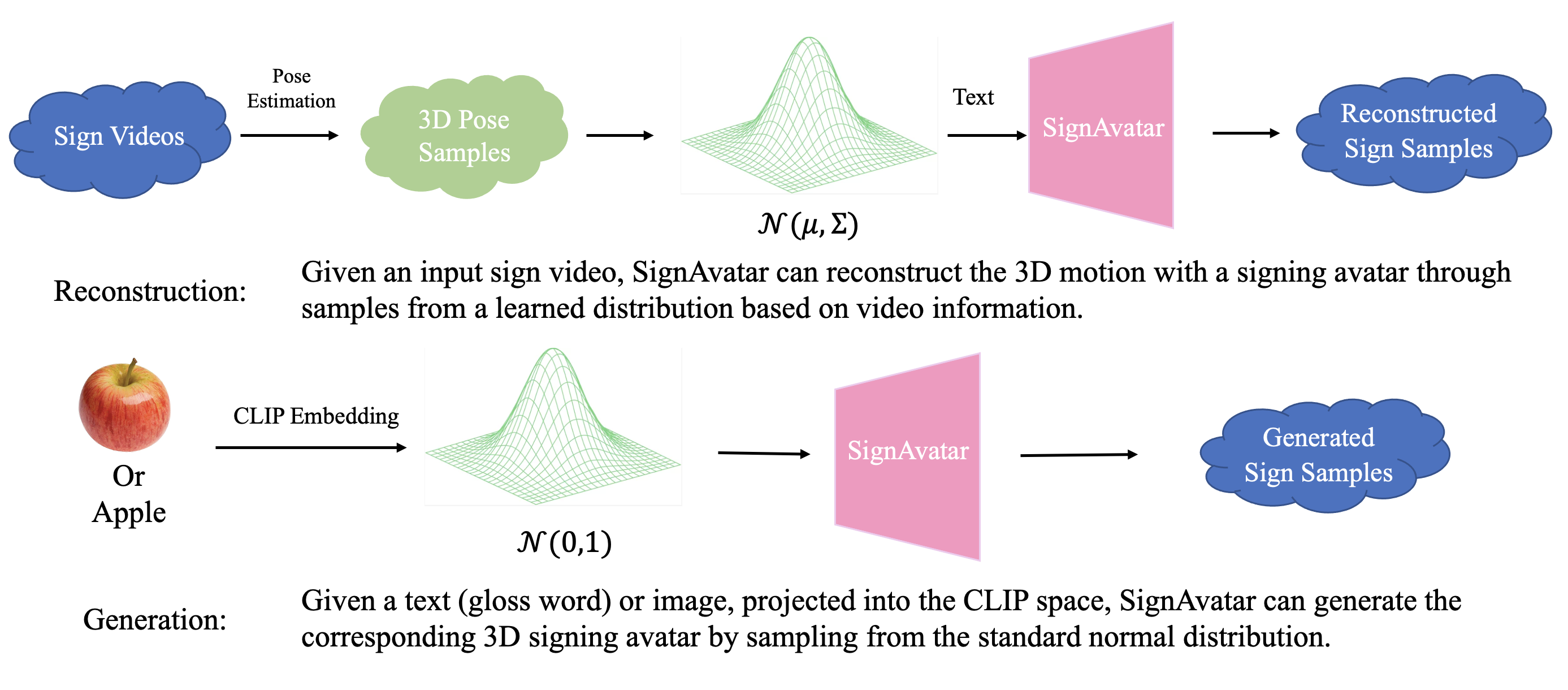}
    \caption{{The upper row represents the reconstruction process, absorbing knowledge and analyzing their relationships, while the lower row indicates the generation process, which outputs knowledge. Unlike the rigid mapping of sign language production, the input text or images in sign language generation showcase greater semantic flexibility.}}
    \label{fig:two_pipeline}
\end{figure*}
An example of this is shown in the middle column of Fig. \ref{fig:upper_body}, where the hand rotation appears to be unrealistic and inaccurate. Since hand pose estimation is crucial for sign language research, we further explored Hand4Whole\cite{moon2022accurate}, which incorporates the metacarpophalangeal (MCP) hand joints to predict 3D wrists. This resulted in more accurate 3D wrist rotation predictions (the last column of Fig. \ref{fig:upper_body}).

We used Hand4Whole to extract SMPL-X parameters from our collection of pre-selected, quality-controlled video frames, thus building the ASL3DWord motion dataset. Meanwhile, we designated the top 30 word samples as the ASL3DWord Subset, which served as the control group for generalization analysis.

\section{Experiments}

\begin{table*}
\caption{ \textbf{Quantitative results comparison on Raw Poses ($Raw$), Reconstructed Poses ($Rec$), and Generated Poses ($Gen$). $\rightarrow$ indicates results are better if they are closer to the extracted Raw pose.}}
\label{table_evaluation}
\begin{center}
\begin{tabular}{|c|c|c|c|c|c|c|c|c|c|}
\hline
\textbf{ASL3DWord Subset} & \textbf{Acc.} $\uparrow$ & \textbf{FID} $\downarrow$ & \textbf{Div.}$\rightarrow$& \textbf{Multi.}$\rightarrow$ & 
{\textbf{ASL3DWord }} & 
{\textbf{Acc.$\uparrow$}}   & 
{\textbf{FID}$\downarrow$}  & 
{\textbf{Div.}$\rightarrow$} & 
{\textbf{Multi.}$\rightarrow$}\\

\hline
$Raw_{train}$  & 1.0 & 0 & 30.001 & 9.921 &  
$Raw_{train}$  & {1.0} & {0} & {34.565} &{ 13.256} \\ 
\hline
$Raw_{test}$ & 0.897 & 0 & 26.252 & 11.180 &  
{$Raw_{test}$} & {0.818} & {0} & {30.599} & {12.289} \\
\hline
\textbf{w/o Curriculum Learning} & & & & &  {\textbf{w/o Curriculum Learning}} & & & &\\
\hline
$Rec_{train}$  & 1.0    & 4.566 & 28.981 & 10.160  &  
{$Rec_{train}$}  & {1.0}  & {3.395} & {33.566} & {13.803} \\
\hline
$Rec_{test}$  & 0.962 & 32.583 & 29.495 & 9.095 &  $Rec_{test}$ & {0.906} & {29.184} & {31.356} & {10.249}\\
\hline
$Gen_{train}$ & 0.884 & 75.243 & 24.566 & 8.250 & $Gen_{train}$ & {0.515 }& {126.830} & {25.732} & {16.500} \\
\hline
$Gen_{test}$  & 0.890 & 65.285 & 24.187 & 6.600 & $Gen_{test}$ & {0.5111} & {100.147} & {25.393} & {12.289}\\
\hline
\textbf{w/ Curriculum Learning} & & & & &  {\textbf{w/ Curriculum Learning}} & & & &\\
\hline
$Rec_{train}$  & 0.997 & 7.195 & 29.340 & 9.993 &  $Rec_{train}$ & {0.999} & {6.112} & {33.583} & {13.347}\\
\hline
$Rec_{test}$  & 0.976 & 32.973 & 29.165 & 7.362 & $Rec_{test}$  & {0.952 }& {40.637} & {32.561} &{ 8.486}\\
\hline
$Gen_{train}$ & 0.946 & 44.469 & 27.115 & 7.160 & $Gen_{train}$ & {0.729} & {85.025} & {27.973 }& {14.700}\\
\hline
$Gen_{test}$  & 0.941 & 46.435 & 27.097 & 5.916 &  $Gen_{test}$  & {0.733} & {71.809} & {27.811} & {10.483} \\
\hline
\end{tabular}
\end{center}
\end{table*}

\begin{table}
\caption{  \textbf{Ablation Study for Quality Control }}
\label{table_quality}
\begin{center}

\begin{tabular}{|c|c|c|c|c|}
\hline
\textbf{Data} & \multicolumn{2}{|c|}{\textbf{w/o Quality Control}} & \multicolumn{2}{|c|}{\textbf{w/ Quality Control}} \\
\hline
\textbf{ASL3DWord Subset} & \textbf{Acc.} $\uparrow$ &  \textbf{FID} $\downarrow$ & \textbf{Acc.} $\uparrow$ &  \textbf{FID} $\downarrow$  \\ 
\hline
$Raw_{train}$  & 1.0 & 0  & 1.0 & 0 \\
\hline
$Raw_{test}$ & 0.790 & 0 & 0.897 &  0\\
\hline
$Rec_{train}$  & 0.927 & 27.319 &0.997 & 7.195 \\
\hline
$Rec_{test}$  & 0.851 & 51.846  & 0.976 & 32.973\\
\hline
$Gen_{train}$ & 0.856 & 70.250  & 0.946 & 44.469\\
\hline
$Gen_{test}$  & 0.860 & 75.515  &0.941 & 46.435\\
\hline
\end{tabular}
\end{center}
\end{table}
\label{sec:experiments}

As depicted in Fig. \ref{fig:two_pipeline}, our SignAvatar can manage both the reconstruction pipeline and generation pipeline. The reconstruction process obtains samples from a distribution learned from the initial extracted poses, which makes it closely resemble the input video. However, the generation process directly samples from a standard normal distribution $N(0, I)$. As a result, the generation process matches the overall motion but exhibits slight differences, such as in the starting and ending positions or the range of motions, as shown in Fig. \ref{fig:quality}. We evaluate the raw, reconstruction, and generation groups using various metrics. First, we introduce the four evaluation metrics used in our experiments. Next, we outline the implementation details. Then, we present an ablation study. Finally, we provide qualitative results.

\subsection{\textbf{Evaluation Metrics} }

We follow the performance measures employed in \cite{guo2020action2motion} for quantitative evaluations.
We measure recognition accuracy, FID, overall diversity, and per-action diversity (referred to as multimodality in \cite{guo2020action2motion}). Below are details of the four metrics:

\begin{itemize}[leftmargin=*,noitemsep,topsep=0pt]
    \item \textbf{Recognition Accuracy (Acc.)} The recognition accuracy metric serves as an indicator of how effectively our reconstruction and generation can be identified by the same classifier. We compute the overall recognition accuracy for three data groups: Raw ($Raw$), Reconstruction ($Rec$), and Generation ($Gen$).
    We begin with a sign motion recognition classifier using Spatio-Temporal Graph Convolutional Networks (STGCN) \cite{yan2018spatial}, trained on $Raw$ data and employing 6D rotations \cite{zhou2019continuity} for pose expression. STGCN seamlessly combines temporal and spatial information, which has recently led to successes in motion recognition tasks.
   
    \item \textbf{Fr\'echet Inception Distance (FID)} FID assesses the overall quality of reconstructed and generated motions by comparing feature distributions. It involves extracting features from the initial raw motions, as well as from the reconstructed and generated motions. Subsequently, FID is computed by measuring the feature distribution of the reconstructed and generated motions against that of the original raw features. This metric holds significant importance and is commonly employed in evaluating the quality of generated motions.

    \item \textbf{Diversity(Div.)} Diversity measures the variance of the motions across all action categories for three data groups. From a set of all motions from various action types within the same data group, two subsets of the same size $S_{d}$ are randomly sampled. Their respective sets of motion feature vectors $\left\{ m_{1},...m_{S_{d}} \right\}$ and $\left\{ m_{1}^{'},...m_{S_{d}}^{'} \right\}$ are extracted. 
    
    The diversity of this set of motions is defined as 
    \begin{equation}\label{eq:example}
        Diversity = \frac{1}{S_{d}}\sum_{i=1}^{D_{d}}\left\| m_{i}-m_{i}^{'} \right\|_{2}\
    \end{equation}
    $S_{d} = 200$ is used in experiments. Hyperparameter settings follow previous motion research \cite{guo2020action2motion, petrovich2021action}.
    
    \item \textbf{Multimoldality(Multi.)} Different from diversity, multimodality measures the average variance within each sign word for three data groups. Given a set of motions with $C$ sign words. For each word, we randomly sample two subsets with same size $S_{l}$ , and then extract two subset of feature vectors $\left\{ m_{c,1},...m_{c,S_{l}} \right\}$ and $\left\{ m_{c,1}^{'},...m_{c,S_{l}}^{'} \right\}$ 
    \begin{equation}
        Multimodality = \frac{1}{C\times S_{l}}\sum_{c=1}^{C}\sum_{i=1}^{S_{l}}\left\| m_{c,i} -m_{c,i}^{'}\right\|_{2}
    \end{equation}
     $C = 30$  and $S_{l} = 20$ are used in experiments following previous motion research.

\end{itemize}

\begin{figure*}[t]
    \centering
    \includegraphics[width=0.9\linewidth]{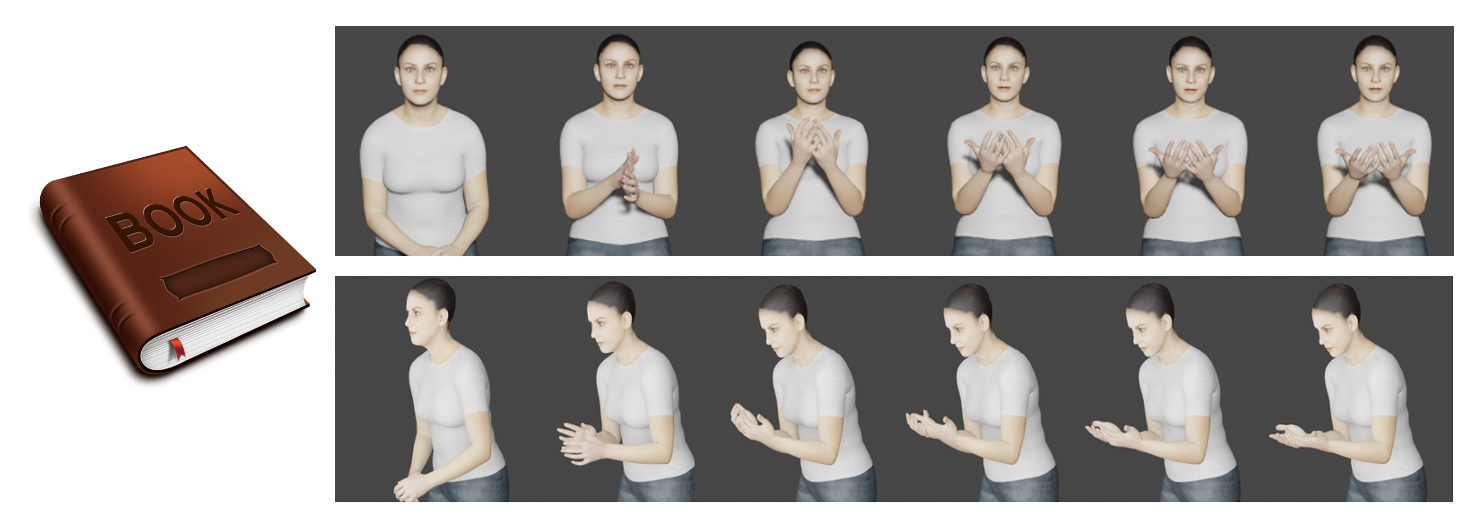}
    \caption{ SignAvatar can accept images as input. Given an image on the left, and using the text-image embedding of CLIP, SignAvatar can recognize the corresponding semantics - "book", and generate the corresponding 3D signing motion. The upper row is the front view and the lower row is the side view.} 
    \label{fig:book}
\end{figure*}

\subsection{\textbf{Ablation Study} }
We perform a comprehensive analysis of the proposed SignAvatar from three perspectives: framework design, curriculum learning strategy, and data collection quality control. We trained and evaluated the proposed model using raw poses ($Raw$), reconstructed poses ($Rec$), and generated poses ($Gen$), while also comparing the individual results of their $train$ and $test$ splits.

\noindent \textbf{SignAvatar CVAE Framework} \quad 
Our experimental results on both the ASL3DWord Subset and ASL3DWord dataset compellingly demonstrate our design's effectiveness in sign motion reconstruction and generation tasks, as shown in Tables~\ref{table_evaluation} and~\ref{table_quality}.
Recognition accuracy serves as a reliable indicator for assessing the quality of reconstructed and generated poses by the SignAvatar CVAE Framework. We train our STGCN-based recognition model using the raw training set.
 From Table~\ref{table_evaluation}, we observe that compared to $Raw_{test}$ accuracy, our CVAE-based Framework significantly improves reconstruction test results, $Rec_{test}$, regardless of whether the curriculum learning strategy is used. 
 This can also be observed in Table ~\ref{table_quality}, where our reconstruction $Rec_{test}$ increases from $0.790$ to $0.851$, even without data collection quality control. This is because the original feature extraction involves inference from images, which are not directly obtained ground truth through camera equipment. When the frame is blurred, such inference can introduce errors and substantial noise. In contrast, our framework possesses a certain denoising effect, resulting in a smoother and more accurate performance.

\noindent \textbf{Curriculum Learning Strategy (CLS)} \quad Through experiments, we have also proved the effectiveness of the CLS for our generation pipeline. Here, recognition accuracy and FID are the two most critical metrics, while diversity and multimodality serve as auxiliary indicators, providing us with insights from the perspective of variance. 
When calculating FID, all data groups are compared with $Raw_{train}$ in the Train Split and with $Raw_{test}$ in the Test Split. Therefore, the values for both $Raw_{train}$ and $Raw_{test}$ are 0.
Analyzing the recognition accuracy from Table \ref{table_evaluation}, when CLS is not used, the value of $Rec_{train}$ reaches 1, but $Rec_{test}$ is 0.962, which is lower than the $Rec_{test}$ value of 0.976 when CLS is employed. Both groups have the same trend, indicating that  CLS improves the model's generalization.
This observation is also well-supported in the generation for different group sizes. Without CLS, the generation performance drops to 0.51, but with CLS, the generation performance remains around 0.73. This indicates CLS improves the model's robustness.
Examining the FID values, it is evident that employing CLS significantly reduces Generation FID. Specifically, for a group word size of 30, $Gen_{train}$ decreases from 75.243 to 44.47, and $Gen_{test}$ from 65.285 to 46.435. For a group word size of 103, it decreases from 100.147 to 71.809.

From the perspectives of Diversity and Multimodality, using CLS results in a slight decrease in \textit{Generation} Multimodality, indicating a sacrifice in within-class variation. However, employing CLS leads to an increase in Diversity, with group sizes increasing from 24 to 27 in the ASL3DWord Subset and from 25 to 27 in the ASL3DWord dataset, bringing the results closer to those of the $Raw$ data group. This suggests that using CLS strikes a balance between different word classes, enhancing overall diversity and making the results more authentic compared to not using it. 
These findings clearly show that our CLS improves the model's robustness, enhances generalization, and boosts authenticity.

\noindent \textbf{Data Collection Quality Control} \quad 
We have also conducted experiments to underscore the importance of quality control during the 3D data collection process. Using a model with CLS and identical configurations, we trained and evaluated the original data for ASL3DWord Subset without quality control. In Table~\ref{table_quality}, we omitted Diversity and Multimodality metrics because their relevance diminishes when the Accuracy and FID results cannot be guaranteed.
The experimental data indicates that $Raw_{test}$ Accuracy is only 79\%, a decrease of 10 percentage points compared to the previous results. This suggests that the original data had a significant amount of noise. Particularly noteworthy is that the accuracy of the generated results is slightly higher than that of the reconstructed results, further confirming the issue of substantial noise in the original data. 
While our SignAvatar enhances performance, leading to improved results for both $Rec_{test}$, $Gen_{train}$ and $Gen_{test}$ compared to $Raw_{test}$, there remains a substantial gap when these are compared to $Rec_{test}$, $Gen_{train}$, and $Gen_{test}$ that have undergone quality control.

Examining the FID metrics, we observe that $Rec_{train}$ increases from 7.195 to 27.319, $Rec_{test}$ increases from 32.973 to 51.846, $Gen_{train}$ increases from 44.47 to 70.250, and $Gen_{test}$ increases from 46.43 to 75.515. This further emphasizes that excessive noise can significantly affect the model's learning ability and generalization performance. Thus, it is evident that quality control of the data is a crucial step in improving data reconstruction and generation quality.

\subsection{\textbf{Qualitative Results} }
We present quality results in Fig. ~\ref{fig:quality}. From the figure, it is evident that the reconstructed results closely match specific provided videos. However, within the same semantic sign language, there may be subtle expressive differences due to individual habits. Our Generation results, on the other hand, reflect this diversity precisely, as they have learned from a vast range of data samples. This illustrates that our SignAvatar can effectively meet the requirements of both reconstruction accuracy and generation diversity. 

SignAvatar can also accept an image as its input condition as shown in Fig. \ref{fig:book}. This is an advantage of using CLIP. With its image encoder and text encoders, CLIP can readily identify the semantic meaning of the image and use this for sign generation from SignAvatar.

\section{CONCLUSIONS AND FUTURE WORKS}
\label{sec:conclusion}

We present SignAvatar, a novel approach for sign language 3D motion reconstruction and generation from 2D isolated videos. 
Our curriculum learning strategy enhances the models' scalability, robustness, generalization, and authenticity. Furthermore, the text-driven and image-driven generation methods increase flexibility in this field. Comprehensive evaluations demonstrate SignAvatar's superior performance in sign language reconstruction and generation tasks. Additionally, we have developed a quality-controlled, SMPL-X-based 3D dataset, ASL3DWord, for academic research.

In the future, we aim to further leverage the semantic space provided by CLIP to explore semantic similarities in sign language. Additionally, recognizing that sign language incorporates non-manual elements such as facial expressions, lip movements, and emotions, we will investigate how facial expressions and body poses contribute to improved understanding, in the context of 3D sign language reconstruction and generation.

\section{ACKNOWLEDGMENTS}

This work is supported in part by the AI Research Institutes program by the National Science Foundation (NSF) and the Institute of Education Sciences (IES), U.S. Department of Education, through Award \# 2229873, NSF Award \# 2223507 and the Institute for Artificial Intelligence and Data Science (IAD) at University at Buffalo. Any opinions,findings and conclusions or recommendations expressed in this material are those of the author(s) and do not necessarily reflect the views of the NFS, the IES, or the IAD.

{
\newpage
\small
\bibliographystyle{ieee}
\bibliography{egbib}
}

\end{document}